\documentclass[twocolumn]{cinc}
\usepackage{graphicx}
\usepackage{comment}

\usepackage{amsmath}
\usepackage{textcomp} 

\usepackage{xcolor}
\usepackage{comment}

\newcommand{\coljma}[1]{{{#1}}}

\usepackage{booktabs}

\usepackage{hyperref}

\usepackage{fancyhdr}
\usepackage{lastpage} 

\begin{document}

\bibliographystyle{cinc}

\title{Electrocardiogram-based diagnosis of liver diseases: an externally validated and explainable machine learning approach}

\author {Juan Miguel Lopez Alcaraz, M.Sc.$^{1}$, Prof. Wilhelm Haverkamp$^{2}$, Prof. Nils Strodthoff$^{1}$$*$ \\
\ \\ 
 $^1$ AI4Health Division, Carl von Ossietzky Universität Oldenburg, Oldenburg, Germany. \\
 \textit{\{juan.lopez.alcaraz,nils.strodthoff\}@uol.de} \\
$^2$ Department of Cardiology, Angiology and Intensive Care Medicine, Charité Campus Mitte, German Heart Center of the Charité-University Medicine, Berlin, Germany. \\
\textit{wilhelm.haverkamp@dhzc-charite.de} \\
$*$ Corresponding author \\  \textit{Accepted version by \textbf{EClinicalMedicine}, The Lancet.}}

\maketitle

\begin{abstract}

\textbf{Background:} \coljma{Liver diseases present a significant global health challenge and often require costly, invasive diagnostics. Electrocardiography (ECG), a widely available and non-invasive tool, can enable the detection of liver disease by capturing cardiovascular-hepatic interactions.}

\textbf{Methods:} \coljma{We trained tree-based machine learning models on ECG features to detect liver diseases using two large datasets: MIMIC-IV-ECG (467,729 patients, 2008–2019) and ECG-View II (775,535 patients, 1994–2013). The task was framed as binary classification, with performance evaluated via the area under the receiver operating characteristic curve (AUROC). To improve interpretability, we applied explainability methods to identify key predictive features.}

\textbf{Findings:} \coljma{The models showed strong predictive performance with good generalizability. For example, AUROCs for alcoholic liver disease (K70) were 0.8025 (95\% confidence interval (CI), 0.8020–0.8035) internally and 0.7644 (95\% CI, 0.7641–0.7649) externally; for hepatic failure (K72), scores were 0.7404 (95\% CI, 0.7389–0.7415) and 0.7498 (95\% CI, 0.7494–0.7509), respectively. The explainability analysis consistently identified age and prolonged QTc intervals (corrected QT, reflecting ventricular repolarization) as key predictors. Features linked to autonomic regulation and electrical conduction abnormalities were also prominent, supporting known cardiovascular-liver connections and suggesting QTc as a potential biomarker.}

\textbf{Interpretation:} \coljma{ECG-based machine learning offers a promising, interpretable approach for liver disease detection, particularly in resource-limited settings. By revealing clinically relevant biomarkers, this method supports non-invasive diagnostics, early detection, and risk stratification prior to targeted clinical assessments.}

\textbf{Funding:} \coljma{This research received no external funding.}

\end{abstract}

\begin{figure*}[!ht]
    \centering
    \includegraphics[width=.8\textwidth]{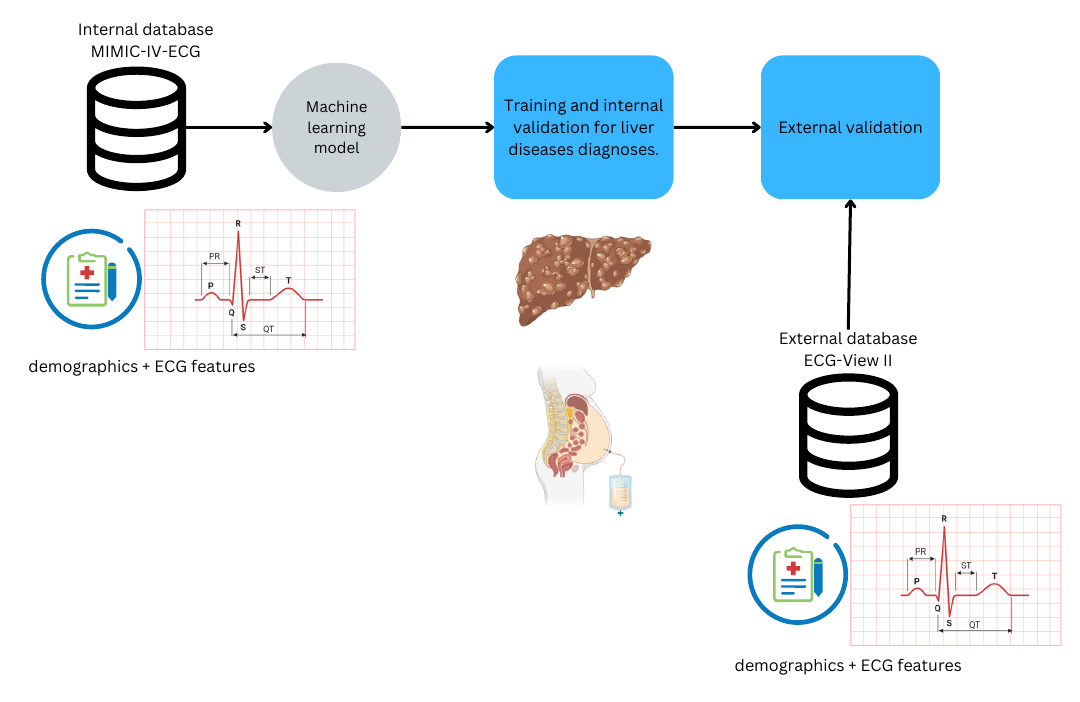}
    \caption{Ilustrive representation of our proposed approach. We use as internal dataset the MIMIC-IV-ECG dataset from which we use as input features demographics and ECG features to train a tree-based model and predict diverse liver diseases. For external validation we take a second cohort of patients from the ECG-View II dataet from which we collect the same ECG features and liver diseases. We define liver diseases by means of ICD10-CM codes for a well define disease representation.}
    \label{fig:abstract}
\end{figure*}

\section*{Research in context}

\subsection*{Evidence before this study}
A systematic search of PubMed, Scopus, and IEEE Xplore (January 2010-December 2024) was conducted using terms like "ECG liver disease diagnoses," and "cardiovascular-liver interaction." Previous research demonstrated the potential of machine learning in healthcare and highlighted the ECG's utility in detecting systemic diseases. However, studies focusing on liver disease detection via ECG were limited, and most lacked the application of explainable AI frameworks or the integration of external validation.

\subsection*{Added value of this study}
This study demonstrates a novel application of ECG data for liver disease detection using explainable machine learning models. By integrating Shapley values, we ensure model transparency, offering insights into feature contributions and their physiological relevance. This approach expands the utility of ECG beyond cardiovascular conditions, filling a critical gap in non-invasive liver disease diagnostics.

\subsection*{Implications of all the available evidence}
Our findings suggest that ECG-based diagnostic tools, combined with interpretable machine learning, have the potential to transform liver disease diagnostics. These tools could facilitate timely detection, improve resource allocation, and inform targeted interventions. Future research should explore integrating ECG data with other non-invasive and low-resource biomarkers such as vital signs and evaluate the generalizability of these methods across diverse populations.

\section{Introduction}

\coljma{Liver diseases represent a significant part of global health challenges, accounting for 2 million deaths annually and responsible for 4\% of all deaths in 2023 \cite{devarbhavi2023global}. Despite the high prevalence of hepatic conditions, timely diagnosis remains a critical challenge. Traditional diagnostic approaches, including blood tests, imaging techniques such as ultrasound, MRI or CT scans, and liver biopsies, are often resource-intensive, invasive, or insufficiently sensitive, particularly in early disease stages \cite{tapper2023diagnoses, ozer2008current, bravo2001liver}. Liver biopsies, although considered the gold standard, carry risks like bleeding and infection, making them unsuitable for routine monitoring. These limitations highlight the urgent need for accessible, non-invasive, and cost-effective diagnostic tools to improve early detection and facilitate timely intervention.}

\coljma{Electrocardiograms (ECG) have long been a cornerstone in diagnosing and monitoring cardiovascular conditions by recording the heart's electrical activity. Traditionally used to detect arrhythmias and myocardial infarction, recent research has broadened their scope to systemic health assessments. ECG-derived models have demonstrated the ability to predict abnormalities in laboratory values \cite{alcaraz2024cardiolablaboratoryvaluesestimation}, patient deterioration \cite{alcaraz2024mds}, and even non-cardiac conditions \cite{topol2021s, siontis2021artificial}. A recent large-scale study further demonstrated the potential of a unified ECG model to predict a wide range of cardiac and non-cardiac conditions \cite{strodthoff2024prospects}. These advances, along with the ECG’s non-invasive nature and widespread availability, position it as a promising tool for innovative diagnostic approaches in systemic diseases, including liver pathologies.}

\coljma{The interplay between liver and cardiovascular health is well-established. Liver diseases often manifest with cardiovascular complications such as cirrhosis-associated cardiomyopathy and portopulmonary hypertension \cite{moller2009cardiovascular}, while cardiac dysfunctions like chronic heart failure can induce hepatic injury, including cardiac-induced liver damage and congestive hepatopathy \cite{naschitz2000heart}. Moreover, shared pathophysiological factors, such as systemic inflammation and electrolyte disturbances like hypokalemia, underscore the bidirectional nature of this relationship \cite{moller2013interactions}.}

\coljma{Parallel to these clinical observations, the application of machine learning (ML) in healthcare has opened new frontiers for disease prediction and diagnosis. While ML models have been applied to liver cancer diagnosis using imaging \cite{yamada2019dynamic} and blood biomarkers \cite{9748503}, most lack external validation and explainability, key elements for clinical trust and adoption \cite{tian2023liver}. Specifically for liver diseases, ECG-based ML has shown some promise; for example, cirrhosis detection using 12-lead ECG and convolutional neural networks (CNN) \cite{ahn2022development}. However, such studies face limitations including black-box modeling, manual class balancing, and absence of external validation. The few works that use more systematic approaches, such as \cite{strodthoff2024prospects}, still lack external validation and often do not explore liver-specific predictive performance.}

\coljma{This study aims to address these gaps by developing a machine learning model that leverages ECG features and demographic data to support the diagnosis of liver diseases. By employing a tree-based approach, we seek to provide a robust, explainable, and externally validated tool that enhances diagnostic capabilities while remaining non-invasive and cost-effective, complementing traditional liver disease diagnostics and advancing clinical decision support systems.}

\section{Methods}

\subsection{Dataset}

\begin{table*}[!ht]
    \centering
    \begin{tabular}{lll}
    \hline
    \textbf{Variable} & \textbf{MIMIC-IV-ECG} & \textbf{ECG-View II} \\ \hline\hline
    \textbf{Gender (\%)} &  &  \\ \hline
    Female     & 226,892 (48.50)  & 375,733 (48.44) \\
    Male   & 240,837 (51.49)  & 399,802 (51.55) \\ \hline
    \textbf{Age (\%)} &  &  \\ \hline
    Median years (IQR) & 66 (25) & 52 (25) \\ 
    Quantile 1 & 18-53 (23.83) & 18-40 (24.03) \\
    Quantile 2 & 53-66 (25.16) & 40-52 (25.75) \\
    Quantile 3 & 66-78 (25.60) & 52-65 (24.94) \\
    Quantile 4 & 78-101 (25.40) & 65-109 (25.28) \\ \hline
    \textbf{ECG features (IQR)} &  &  \\ \hline
    RR-interval & 769 (264)  & 857 (227) \\
    PR-interval & 158 (38)  & 158 (28) \\
    QRS-duration & 94 (23)  & 90 (14) \\
    QT-interval & 394 (68)  & 392 (48) \\
    QTc-interval & 447 (47)  & 421 (37) \\
    P-wave axis & 51 (32) & 53 (28) \\
    QRS-axis & 13 (61) & 48 (49) \\
    T-wave axis & 42 (58)  & 44 (33) \\ \hline
    \textbf{Number of samples (unique patients) with condition} & & \\ \hline
    K70: Alcoholic liver disease &   10333(2428)  &  5194(2533)  \\
    K703: Alcoholic cirrhosis of liver &   8786(1951)   & 2185(908)   \\
    K7030: Alcoholic cirrhosis of liver without ascites & 6159(1592)  & 279(93)  \\
    K72:Hepatic failure, not elsewere classified & 8357(1974)  & 736(299)  \\
    K729: Hepatic failure, unspecified &  5163(1384) &  736(299)  \\
    K7290: Hepatic failure, unspecified without coma & 3111(863)  & 237(73) \\ \hline
    \end{tabular}
    \caption{Summary of variables characteristics across samples including demographics such as gender counts (ratio), age median in years (IQR), as well as age distribution by quantiles (ratio). Similarly the median (IQR) of ECG features such as RR-interval, PR-interval, QRS-duration, QT-interval, QTc-interval in milliseconds, P-wave axis, QRS-axis, as well as T-wave axis in degrees. Finally, we have included the number of positive samples for the investigated conditions across each dataset and the number of unique patients representing that population in brackets.}
    \label{tab:descriptive}
\end{table*}

Our primary dataset for training and internal evaluation comes from the MIMIC-IV-ECG database \cite{johnson2023mimic,MIMICIVECG2023} which is a dataset from patients admitted to the Beth Israel Deaconess Medical Center in Boston, Massachusetts, where the population represents admission at the emergency department (ED) and intensive care unit (ICU). We define as target variables discharge diagnoses coded using the International Classification of Diseases Clinical Modification (ICD-10-CM). The investigated codes come from ICD-chapter XI covering diseases of the digestive system: K70, K703, and K7030, for alcoholic liver disease and sub-conditions, as well as K72, K729, and K7290 for hepatic failure and sub-conditions. To build a comprehensive feature set, we align the ECG features from MIMIC-IV to those from the ECG-VIEW-II database \cite{kim2017ecg}, our secondary dataset used for external validation, and is a dataset from patients admitted to a South Korean tertiary teaching hospital. The combined feature set includes: ECG metrics (RR interval, PR interval, QRS duration, QT interval, QTc interval in milliseconds; P wave axis, QRS axis, T wave axis in degrees) and basic patient demographics (sex as binary and age as a continuous variable). We carefully applied Bazett’s formula to compute the QTc interval in MIMIC-IV, ensuring consistency with the QTc calculation used in ECG-VIEW-II. We did not apply any imputation into the dataset due to the inherently nature of the utilized model to handle missing data. For the internal dataset, we use stratified folds based on diagnoses, age, and gender distributions, following an 18:1:1 split as described in previous work \cite{strodthoff2024prospects}. For the external dataset, we apply a similar stratification procedure to ensure consistency. We point out that we ensure no patient overlap across folds but rather balanced distribution of age, gender, and diagnoses. We adopt the corresponding order of datasets for internal training and external validation regardless of the sample size given that MIMIC-IV-ECG provides more ethnically diverse data than ECG-View II, enhancing learning across diverse distributions as previously seen in \cite{alcaraz2024estimationcardiacnoncardiacdiagnosis}, which employs a similar approach but does not cover liver diseases. Our final datasets consist of two large cohorts of 467,729 samples for training and 775,535 samples for external validation. \coljma{Race data were not collected from the MIMIC-IV-ECG or ECG-VIEW-II datasets for this study, as the focus was on generalization across diverse populations, see Table \ref{tab:descriptive} for a detailed summary of variables and label distribution of both datasets, and figure \ref{fig:abstract} for an ilustration of the pipeline used in this work. We provide a TRIPOD checklist related to this study in the supplementary material.}

\subsection{Models and evaluation}
In this study, we develop individual tree-based models using Extreme Gradient Boosting (XGBoost) to solve binary classification tasks, one for each considered ICD-10-CM code. During training, early stopping is applied with a patience of 10 iterations on the validation fold to prevent overfitting. Model performance is primarily assessed using the area under the receiver operating characteristic curve (AUROC) on the validation fold, ensuring robust model selection. We do not implement any imbalance correction methods as they tend to negatively impact model calibration \cite{Carriero2025}. To evaluate generalization, we report AUROC scores on both the internal test set and the external dataset, along with corresponding 95\% prediction intervals derived from empirical bootstrapping using 1000 iterations. Additionally, we assess model performance using sensitivity and specificity at a fixed sensitivity threshold of 0.70. This approach ensures a consistent evaluation framework across datasets, allowing for direct comparison of specificity while maintaining a clinically relevant sensitivity level. Finally, we report the prevalence of each condition in both datasets, providing context for model performance in relation to class distribution. Please refer to supplementary material for the list of model hyperparameters and a short discussion on the matter.

We emphasize that gradient-boosted decision trees such as XGBoost models represent a well-accepted model choice for structured tabular as considered in this work \cite{grinsztajn2022tree,mcelfresh2023neural,shwartz2022tabular}. Currently, deep learning methods do not reach their performance levels, in particular in the limit of large sample sizes \cite{zabergja2024tabular}. Deep learning models applied directly to raw ECG waveforms, as in \cite{strodthoff2024prospects}, might achieve superior performance in most scenarios, but this is not guaranteed \cite{ott2024using}. ECG features possess unique advantages in terms of interpretability, robustness, and a simpler deployment.

\subsection{Explainability}
We aim to go beyond a performance evaluation by providing insights into the trained models. To this end, we integrate Shapley values into our pipeline \cite{lundberg2020local}. Shapley values provide a measure of feature importance by quantifying how much each feature contributes to the model prediction.

\subsection{Ethics}
\coljma{Both datasets used in this study, MIMIC-IV and ECG-VIEW-II, are publicly available and fully de-identified in accordance with the US Health Insurance Portability and Accountability Act (HIPAA) privacy rule. MIMIC-IV’s release is overseen by the Institutional Review Boards (IRBs) of Beth Israel Deaconess Medical Center and the Massachusetts Institute of Technology. ECG-VIEW-II follows similar de-identification standards, excluding unique identifiers and applying additional privacy measures, such as top-coding extreme laboratory values, removing highly stigmatized diagnoses, obfuscating date information within a randomized range while preserving relative intervals, and grouping birth years into five-year intervals. As both datasets are anonymized, no additional ethical approval was required for this study. The mode of consent for both datasets is implied consent, as they are publicly available and de-identified, ensuring that individual patient identities are not accessible, in compliance with privacy regulations.}

\subsection{Data and code availability} 
\coljma{The code for dataset preprocessing and experimental replications can be found in a dedicated GitHub repository: https://github.com/AI4HealthUOL/CardioDiag.}

\subsection{Role of the funding source}
\coljma{No funding was received for this study. The authors were not paid to write this article by any company or other agency. The corresponding author affirms that all authors had full access to all the data in the study and accept responsibility for the decision to submit for publication.}

\section{Results}

\subsection{Predictive performance}

\begin{table*}[ht!]
    \centering
    \begin{tabular}{lcccc}
    \hline
    \textbf{Condition} & \textbf{AUROC} & \textbf{Sensitivity} & \textbf{Specificity} & \textbf{Prevalence} \\ \hline\hline
    \multicolumn{5}{c}{\textbf{MIMIC-IV-ECG (Internal)}} \\ \hline
    K70: Alcoholic liver disease & 0.8025 (0.8020, 0.8035) & 0.7012 & 0.7361 & 2.21\% \\
    K703: Alcoholic cirrhosis of liver & 0.7887 (0.7880, 0.7887) & 0.7018 & 0.7136 & 1.88\% \\
    K7030: Alcoholic cirrhosis of liver without ascites & 0.7819 (0.7805, 0.7817) & 0.7013 & 0.7177 & 1.32\% \\
    K72: Hepatic failure, not elsewhere classified & 0.7404 (0.7389, 0.7415) & 0.7015 & 0.6871 & 1.79\% \\
    K729: Hepatic failure, unspecified & 0.7647 (0.7631, 0.7647) & 0.7031 & 0.7064 & 1.1\% \\
    K7290: Hepatic failure, unspecified without coma & 0.7833 (0.7817, 0.7851) & 0.7047 & 0.7151 & 0.67\% \\ \hline
    \multicolumn{5}{c}{\textbf{ECG-View II (External)}} \\ \hline
    K70: Alcoholic liver disease &  0.7644 (0.7641, 0.7649) & 0.7000 & 0.6762 & 0.67\% \\
    K703: Alcoholic cirrhosis of liver & 0.8590 (0.8589, 0.8594) & 0.7002 & 0.8379 & 0.28\% \\
    K7030: Alcoholic cirrhosis of liver without ascites & 0.8777 (0.8770, 0.8783) & 0.7025 & 0.8553 & 0.04\% \\
    K72: Hepatic failure, not elsewhere classified & 0.7498 (0.7494, 0.7509) & 0.7011 & 0.6831 & 0.09\% \\
    K729: Hepatic failure, unspecified & 0.7821 (0.7814, 0.7823) & 0.7011 & 0.7360 & 0.03\% \\
    K7290: Hepatic failure, unspecified without coma & 0.8003 (0.7988, 0.8011) & 0.7215 & 0.7532 & 0.09\% \\ \hline
    \end{tabular}
    \caption{Comparison of AUROC, sensitivity, specificity, and prevalence for each condition in the internal (MIMIC-IV-ECG) and external (ECG-View II) datasets. sensitivity and specificity values were measured at a fixed sensitivity threshold of 0.70.}
    \label{tab:comparison}
\end{table*}

Table \ref{tab:comparison} illustrates the predictive performance of our model across multiple liver conditions, assessed using AUROC scores, sensitivity and specificity on both internal and external test sets. The reported 95\% AUROC prediction intervals provide insights into the reliability of these estimates. The reported prevalences of each condition within the respective datasets highlight substantial differences between the two cohorts. The MIMIC-IV-ECG cohort exhibits prevalence rates between 0.67\% and 2.21\%, whereas the ECG-View II cohort demonstrates markedly lower prevalences, ranging from 0.03\% to 0.67\%. Overall, alcoholic liver disease (K70) and its subcategories (K703, K7030) demonstrate consistently strong discriminative performance, with AUROC values ranging from 0.78 to 0.80 in the internal dataset and improving to 0.86–0.88 in the external dataset. This suggests that ECG-derived features generalize well for detecting these conditions despite variations in dataset composition. In contrast, hepatic failure (K72) and its subcategories (K729, K7290) show lower AUROC values, particularly in the internal dataset, where they range from 0.74 to 0.78. External validation results for these conditions remain comparable but exhibit greater variability.

\subsection{Explainability}

\begin{figure*}[ht!]
    \centering
    \includegraphics[width=\textwidth]{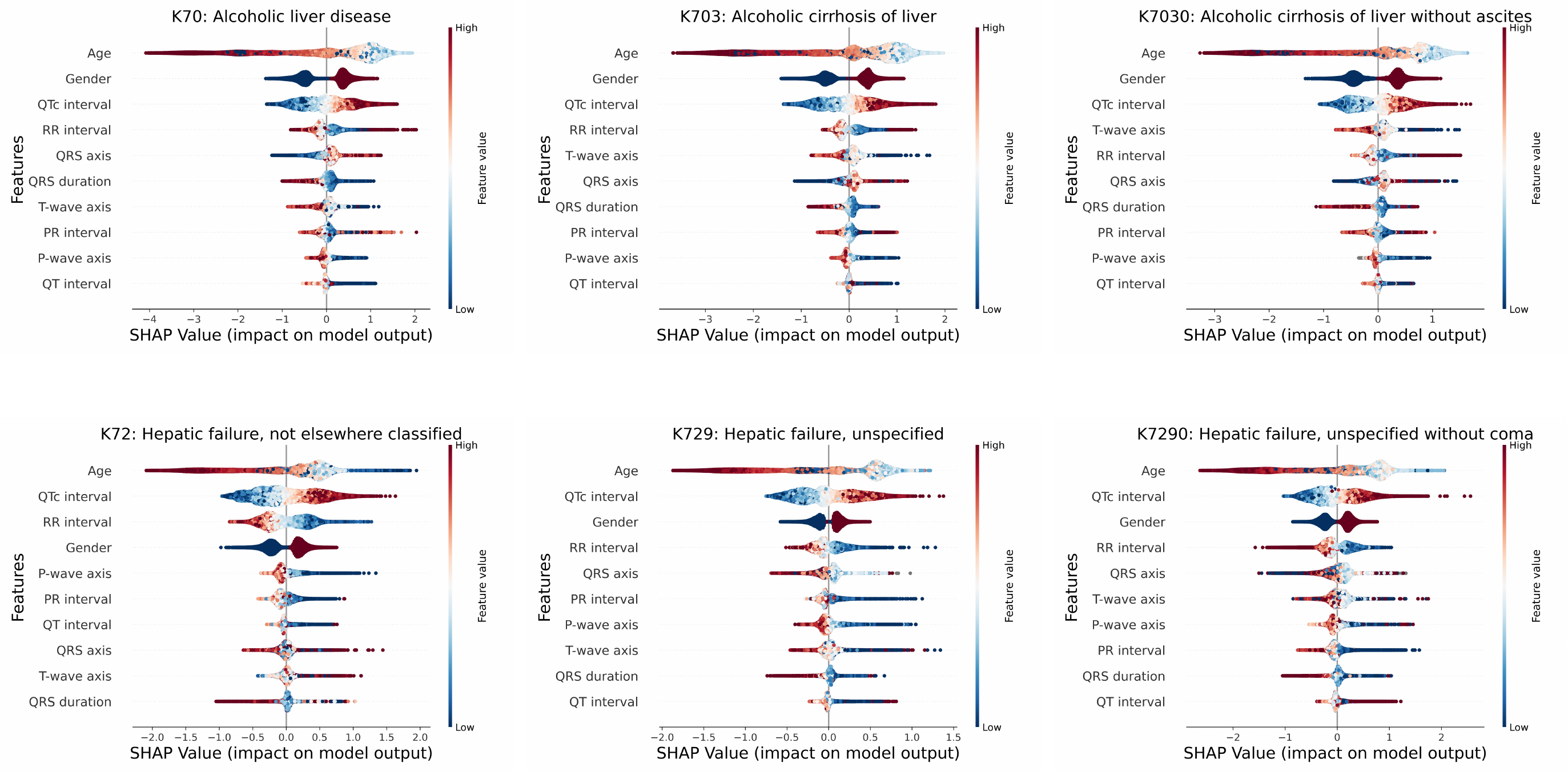}
    \caption{Explainability results for the six investigated conditions. The beeswarm plot indicates for every sample and every feature if the corresponding feature contributes positively (right hand side) or negatively (left hand side) to the model prediction. An additional color-coding indicates if a data point is associated with high (red) or low (blue) feature values.}
    \label{fig:shapley}
\end{figure*}

Figure \ref{fig:shapley} presents the results of the explainability analysis based on Shapley values. 
Across all conditions, age emerges as the most important predictor, with a consistent negative effect for both high and low ages, while intermediate ages have a positive effect, except in the case of ``hepatic failure not elsewhere classified'', where low age contributes positively. Gender also plays a significant role, with males being the predominant group contributing most. Additionally, high QTc values, which are the most important ECG feature across all conditions, further influence the outcomes.

In alcoholic liver diseases and their sub-conditions, a consistent ranking is observed across all three conditions, with age, gender, QTc, and RR interval having effects as described above, with the exception that T-wave axis features are more important than RR intervals in cases of cirrhosis without ascites. The RR interval shows that both low and very high values contribute positively, with the latter having an even stronger effect. T-wave shows that low values contribute positively for all conditions. The QRS axis predominantly shows high values, although some low values are noted in cases without ascites. Similarly, the QRS duration tends to be low, but higher values are also seen in cases without ascites.

In hepatic failure and its sub-conditions, a consistent ranking is observed across all three conditions, with age, QTc, gender, and RR interval being the top four predictors, where age, gender, and QTc have effects as described above. For the RR interval, low values contribute positively, while high values have a negative impact across all conditions. The QRS axis primarily shows positive contributions from median values in hepatic failure, with high values having a negative effect. In contrast, for hepatic failure without coma, median values still contribute positively, but low values have a negative effect. Regarding the PR interval, median values contribute negatively in hepatic failure, while high values have a negative effect in cases without coma, with low values contributing positively for both. Finally, the P-wave axis shows that low values contribute positively for hepatic failure, with high values contributing negatively, whereas in hepatic failure without coma, some very high values contribute positively.

\section{Discussion}

Detecting liver diseases through ECG features may initially seem unconventional, as the ECG is traditionally associated with diagnosing cardiovascular conditions. The connection between the heart and liver is less apparant but underscores the importance of biomarkers in diagnostic innovations. While the physiological mechanisms linking liver dysfunction to ECG abnormalities are not yet fully understood, they represent an intriguing area for further research. Our findings reveal specific patterns in the ECG data that act as a distinctive fingerprint for liver diseases. These patterns suggest potential physiological interactions between the heart and liver, detectable through the applied methods. This novel concept highlights the promise of interdisciplinary approaches, bridging cardiology and hepatology, to uncover new diagnostic pathways. The current work could help to establish the ECG as a screening tool prior to targeted clinical assessment, supporting early detection and risk stratification.

The exceptional predictive power of a small set of ECG features highlights their ability to detect liver conditions from a single ECG with high accuracy. Robust AUROC values across internal and external validations demonstrate the reliability of these features, even in diverse cohorts. The distinct patterns observed for conditions like alcoholic liver disease and hepatic failure underscore the physiological links between cardiac and hepatic health.

Previous research identified ascites as a significant confounding factor in detecting cirrhosis from ECG \cite{OuyangMICCAI23}. At this point, it is important to stress a difference in the study design compared to the conventional control-group approach that was first pointed out in \cite{strodthoff2024prospects}. Rather than selecting a control group that matches the diseased subgroup according to as many covariates as possible, we use the remainder of the entire study cohorts not associated with the condition under consideration as negatives, which represents the most realistic negative control group conceivable, see discussion in the the literature on the challenges of control group matching \cite{stuart2010matching,king2019propensity}.

Previous analysis \cite{strodthoff2024prospects} explicitly examined label overlaps and confirmed that the MIMIC-IV dataset is large and diverse enough to minimize substantial confounding due to co-occurring cardiovascular diseases. This suggests that the model is not merely detecting cardiovascular conditions but capturing meaningful ECG signatures associated with liver pathology. However, other potential confounders, such as overall disease severity, mortality risk, or medication effects, may still influence predictions, which lay ground for future work.

In addition to externally validated conditions, we provide a broader assessment of the model’s predictive performance across various hepatic and gastrointestinal diseases, as our analysis focused on the ICD-10 K00–K95 block. Table \ref{tab:icd10_codes} (Supplementary material) includes both chronic and acute conditions, such as acute alcoholic pancreatitis (K852), acute and subacute hepatic failure (K720, K7200), and acute appendicitis (K3589), offering further insight into the model’s applicability beyond long-term liver disease.

To address concerns regarding the influence of demographic features, we compared models trained on ECG features alone versus demographic data alone Table \ref{tab:ecg_demographics} (Supplementary material). This analysis demonstrates that while demographic factors are more predictive for alcoholic liver disease, ECG signals are stronger indicators of hepatic failure, suggesting that ECG-based predictions capture relevant physiological alterations beyond population-level associations. The models operating on the combined feature sets show in all cases a statistically significant improvement over single-modality models, which stresses the complementarity of both feature sets.

To evaluate the robustness of our findings, we analyzed model performance within different demographic subgroups, defined by gender and age quantiles, as detailed in Table \ref{tab:demographic_subgroups} (Supplementary Material). The AUROC values remain stable across most groups, though a performance decline is observed in the oldest age quantile. Notably, female patients generally exhibit higher AUROC values than male patients, even when slightly underrepresented. This suggests potential pathophysiological differences influencing model predictability. These findings indicate that the model effectively captures ECG signatures of liver disease while maintaining generalizability across diverse demographic subgroups.


The significance of aging is certainly an expected outcome, as it has been previously proved that eldery have more agreement with the inflamm-aging theory, in which aging accrues inflammation \cite{sheedfar2013liver}.
The prevalence of hepatocellular carcinoma (HCC) is higher in men, possibly due to differences in risk factor exposure \cite{hashimoto2011prevalence}. This is consistent with findings in cirrhosis and liver failure, where men are more frequently affected due to the progression of liver disease towards HCC. The study also suggests that estrogen may influence HCC pathogenesis, highlighting potential gender differences in the progression of liver disease and its complications, including liver failure.

Previous research \cite{zhao2016prevalence} highlighted the prevalence of prolonged (high) QTc values in alcoholic cirrhosis within an Asian cohort. A second study \cite{bal2003prolongation} found that liver transplantation significantly improved QTc values in about half of the patients, suggesting that liver disease contributes to prolonged QTc but may not be the only factor involved. Our study expands on these findings by examining a diverse cohort across different ethnicities and by including not only cirrhosis but also hepatic failure cases. It supports the idea that while liver disease is a key contributor, the pathogenesis of prolonged QTc is multifactorial. A study by Toma et al \cite{toma2020electrocardiographic} found an accentuated deceleration of the T-wave in cirrhotic patients, indicating repolarization abnormalities. Similarly, we observed low T-wave axis values in cirrhosis, confirming this altered repolarization. However, in liver failure complications, we found a novel increase in T-wave axis values, suggesting a shift in repolarization patterns as the disease progresses. Similarly, another study \cite{lee2022prolonged} suggested longer RR intervals in cirrhotic patients, here we also confirm the finding on very high values, however, as the T-wave, RR intervals shift directions on complications such as liver failure as low values contribute most positively. To summarize, we observed a shift in T-wave axis values and RR intervals between cirrhotic patients and those with liver failure complications, highlighting evolving repolarization patterns and potential alterations in cardiac rhythm as the disease progresses.

Firstly, the most promising and direct application would be the development of an externally validated unified AI model that can simultaneously assess liver and heart conditions using the shared systemic effects visible in ECG signals. This approach not only improves diagnostic accuracy, but also reduces the need for multiple separate diagnostic tests, potentially streamlining patient evaluation in busy clinical settings. Secondly, by detecting subtle changes in the ECG that signal early systemic disturbances, such as fluid overload or electrolyte imbalances, the model could enable earlier interventions, which are critical for improving patient outcomes, particularly in cases of liver disease and its complications. Lastly, another application is the ability of the model to guide diagnostic pathways by detecting ECG abnormalities that suggest underlying liver or myocardial dysfunction, which could prompt clinicians to pursue additional investigations, such as imaging or laboratory tests, that would otherwise have been delayed or overlooked. In this way, the model acts as a clinical decision support tool that enhances the diagnostic process, improving both efficiency and accuracy. By integrating insights both from the cardiac and the hepatic systems, this approach could ultimately improve patient monitoring, facilitate earlier disease detection, and reduce healthcare costs associated with misdiagnoses or delayed treatment.


While our study highlights the potential of ECGs in detecting, as well as potentially screening and monitoring liver conditions, future work could focus on how ECG abnormalities vary across age groups and how these variations differ from healthy aging features \cite{ott2024using}. Also beyond identifying associations, future research could aim to explore causal relationships between ECG patterns and digestive and liver conditions along the lines of \cite{alcaraz2024causalconceptts}. Additionally, an in-depth study on liver diseases based on raw waveforms would be in order extending \cite{strodthoff2024prospects,alcaraz2024mds} with proper external validation. In the light of the recent findings on the superiority of raw ECG waveforms compared to ECG features for diagnostic tasks \cite{alcaraz2024mds}, further improvements in diagnostic accuracy are expected in this scenario. Future work should further explore possible confounding factors such as overall disease severity, mortality risk, or medication effects, to refine the physiological interpretation of ECG alterations in liver disease. In this respect, it would be insightful extend the modeling scheme beyond binary conditions to multiple output classes in order to gain insights into the effect of ECG features on disease severity. Finally, while ICD-10-CM codes provide a standardized diagnostic scheme, their primary use in billing may introduce biases across institutions. We acknowledge this limitation but note that no large datasets with gold-standard diagnostics, such as biopsies or imaging, currently exist. Given our goal of broad applicability, ICD codes remain a valuable proxy for large-scale clinical studies, though future work should explore more clinically reliable methods for the definition of target variables.

\section*{Contribution}
JMLA and NS conceived and designed the project (Conceptualization, Methodology). JMLA conducted all experimental analyses (Data Curation, Formal Analysis, Software, Investigation), with NS providing supervision (Supervision, Validation) and WH contributing critical revision of the clinical content (Validation, Resources). JMLA wrote the original draft (Writing – Original Draft), while NS and WH revised it (Writing – Review \& Editing). All authors accessed and verified the underlying data, critically revised the manuscript, and approved the final version for publication.

\section*{Data sharing}
\coljma{The datasets used in this study are publicly available, but we do not share the data directly. Researchers interested in accessing the deidentified participant data and the corresponding data dictionary should request it from the original data sources. We provide only the code used to process the experiments. The MIMIC-IV dataset can be accessed through https://doi.org/10.13026/4nqg-sb35. For the ECG-VIEW-II dataset, access can be obtained upon request through the institutional contacts, and the related publication is available via https://doi.org/10.1371/journal.pone.0176222. Related documents, including the study protocol and statistical analysis plan, are available on reasonable request.}

\section*{Declaration of interests}
The authors declare no competing interests.

\bibliography{refs}

\begin{thebibliography}{10}
\expandafter\ifx\csname url\endcsname\relax
  \def\url#1{\texttt{#1}}\fi
\expandafter\ifx\csname urlprefix\endcsname\relax\def\urlprefix{URL }\fi

\bibitem{devarbhavi2023global}
Devarbhavi H, Asrani SK, Arab JP, Nartey YA, Pose E, Kamath PS.
\newblock Global burden of liver disease: 2023 update.
\newblock Journal of hepatology 2023;\hspace{0pt}79(2):516--537.
\newblock Doi: \url{https://doi.org/10.1016/j.jhep.2023.03.017}.

\bibitem{tapper2023diagnoses}
Tapper EB, Parikh ND.
\newblock Diagnosis and management of cirrhosis and its complications: a review.
\newblock Jama 2023;\hspace{0pt}329(18):1589--1602.
\newblock Doi: \url{https://doi.org/10.1001/jama.2023.5997}.

\bibitem{ozer2008current}
Ozer J, Ratner M, Shaw M, Bailey W, Schomaker S.
\newblock The current state of serum biomarkers of hepatotoxicity.
\newblock Toxicology 2008;\hspace{0pt}245(3):194--205.
\newblock Doi: \url{https://doi.org/10.1016/j.tox.2007.11.021}.

\bibitem{bravo2001liver}
Bravo AA, Sheth SG, Chopra S.
\newblock Liver biopsy.
\newblock New England Journal of Medicine 2001;\hspace{0pt}344(7):495--500.
\newblock Doi: \url{https://doi.org/10.1056/NEJM200102153440706}.

\bibitem{alcaraz2024cardiolablaboratoryvaluesestimation}
Alcaraz JML, Strodthoff N.
\newblock Cardiolab: Laboratory values estimation and monitoring from electrocardiogram signals -- a multimodal deep learning approach.
\newblock arXiv preprint arXiv241114886 2024;\hspace{0pt}Doi: \url{https://arxiv.org/abs/2411.14886}.

\bibitem{alcaraz2024mds}
Alcaraz JML, Strodthoff N.
\newblock Mds-ed: Multimodal decision support in the emergency department--a benchmark dataset for diagnoses and deterioration prediction in emergency medicine.
\newblock arXiv preprint arXiv240717856 2024;\hspace{0pt}Doi: \url{https://arxiv.org/abs/2407.17856}.

\bibitem{topol2021s}
Topol EJ.
\newblock What's lurking in your electrocardiogram?
\newblock The Lancet 2021;\hspace{0pt}397(10276):785.
\newblock Doi: \url{https://doi.org/10.1016/S0140-6736(21)00452-9}.

\bibitem{siontis2021artificial}
Siontis KC, Noseworthy PA, Attia ZI, Friedman PA.
\newblock Artificial intelligence-enhanced electrocardiography in cardiovascular disease management.
\newblock Nature Reviews Cardiology 2021;\hspace{0pt}18(7):465--478.
\newblock Doi: \url{https://doi.org/10.1038/s41569-020-00503-2}.

\bibitem{strodthoff2024prospects}
Strodthoff N, Lopez~Alcaraz JM, Haverkamp W.
\newblock Prospects for artificial intelligence-enhanced electrocardiogram as a unified screening tool for cardiac and non-cardiac conditions: an explorative study in emergency care.
\newblock European Heart Journal Digital Health 2024;\hspace{0pt}ztae039.
\newblock Doi: \url{https://doi.org/10.1093/ehjdh/ztae039}.

\bibitem{moller2009cardiovascular}
M{\o}ller S, Henriksen JH.
\newblock Cardiovascular complications of cirrhosis.
\newblock Postgraduate medical journal 2009;\hspace{0pt}85(999):44--54.
\newblock Doi: \url{https://doi.org/10.1136/gut.2006.112177}.

\bibitem{naschitz2000heart}
Naschitz JE, Slobodin G, Lewis RJ, Zuckerman E, Yeshurun D.
\newblock Heart diseases affecting the liver and liver diseases affecting the heart.
\newblock American heart journal 2000;\hspace{0pt}140(1):111--120.
\newblock Doi: \url{https://doi.org/10.1067/mhj.2000.107177}.

\bibitem{moller2013interactions}
M{\o}ller S, Bernardi M.
\newblock Interactions of the heart and the liver.
\newblock European heart journal 2013;\hspace{0pt}34(36):2804--2811.
\newblock Doi: \url{https://doi.org/10.1093/eurheartj/eht246}.

\bibitem{yamada2019dynamic}
Yamada A, Oyama K, Fujita S, Yoshizawa E, Ichinohe F, Komatsu D, Fujinaga Y.
\newblock Dynamic contrast-enhanced computed tomography diagnosis of primary liver cancers using transfer learning of pretrained convolutional neural networks: Is registration of multiphasic images necessary?
\newblock International Journal of Computer Assisted Radiology and Surgery 2019;\hspace{0pt}14:1295--1301.
\newblock Doi: \url{https://doi.org/10.1007/s11548-019-01987-1}.

\bibitem{9748503}
Kim Y, Kim J, Lee S, Ahn S, Kim J, Park S, Cho S.
\newblock Automatic hepatocellular carcinoma diagnosis using graph convolutional network.
\newblock In 2022 International Conference on Electronics, Information, and Communication (ICEIC). 2022;\hspace{0pt} 1--4.
\newblock Doi: \url{https://doi.org/10.1109/ICEIC54506.2022.9748503}.

\bibitem{tian2023liver}
Tian Y, Liu M, Sun Y, Fu S.
\newblock When liver disease diagnosis encounters deep learning: Analysis, challenges, and prospects.
\newblock iLIVER 2023;\hspace{0pt}2(1):73--87.
\newblock Doi: \url{https://doi.org/10.1016/j.iliver.2023.02.002}.

\bibitem{ahn2022development}
Ahn JC, Attia ZI, Rattan P, Mullan AF, Buryska S, Allen AM, Kamath PS, Friedman PA, Shah VH, Noseworthy PA, et~al.
\newblock Development of the ai-cirrhosis-ecg score: an electrocardiogram-based deep learning model in cirrhosis.
\newblock Official journal of the American College of Gastroenterology ACG 2022;\hspace{0pt}117(3):424--432.
\newblock Doi: \url{https://doi.org/10.14309/ajg.0000000000001617}.

\bibitem{johnson2023mimic}
Johnson AE, Bulgarelli L, Shen L, Gayles A, Shammout A, Horng S, Pollard TJ, Hao S, Moody B, Gow B, et~al.
\newblock Mimic-iv, a freely accessible electronic health record dataset.
\newblock Scientific data 2023;\hspace{0pt}10(1):1.
\newblock Doi: \url{https://doi.org/10.1038/s41597-023-02136-9}.

\bibitem{MIMICIVECG2023}
Gow B, Pollard T, Nathanson LA, Johnson A, Moody B, Fernandes C, Greenbaum N, Waks JW, Eslami P, Carbonati T, Chaudhari A, Herbst E, Moukheiber D, Berkowitz S, Mark R, Horng S.
\newblock Mimic-iv-ecg: Diagnostic electrocardiogram matched subset, 2023.
\newblock Doi: \url{https://doi.org/10.13026/4nqg-sb35}.

\bibitem{kim2017ecg}
Kim YG, Shin D, Park MY, Lee S, Jeon MS, Yoon D, Park RW.
\newblock Ecg-view ii, a freely accessible electrocardiogram database.
\newblock PloS one 2017;\hspace{0pt}12(4):e0176222.
\newblock Doi: \url{https://doi.org/10.1371/journal.pone.0176222}.

\bibitem{alcaraz2024estimationcardiacnoncardiacdiagnosis}
Alcaraz JML, Strodthoff N.
\newblock Estimation of cardiac and non-cardiac diagnosis from electrocardiogram features.
\newblock In 2024 Computing in Cardiology Conference (CinC). IEEE, 2024;\hspace{0pt} 1--4.
\newblock Doi: \url{https://arxiv.org/abs/2408.17329}.

\bibitem{Carriero2025}
Carriero A, Luijken K, de~Hond A, Moons KGM, van Calster B, van Smeden M.
\newblock The harms of class imbalance corrections for machine learning based prediction models: A simulation study.
\newblock Statistics in Medicine January 2025;\hspace{0pt}44(3–4).
\newblock ISSN 1097-0258.
\newblock \urlprefix\url{http://dx.doi.org/10.1002/sim.10320}.

\bibitem{grinsztajn2022tree}
Grinsztajn L, Oyallon E, Varoquaux G.
\newblock Why do tree-based models still outperform deep learning on typical tabular data?
\newblock Advances in neural information processing systems 2022;\hspace{0pt}35:507--520.

\bibitem{mcelfresh2023neural}
McElfresh D, Khandagale S, Valverde J, Prasad~C V, Ramakrishnan G, Goldblum M, White C.
\newblock When do neural nets outperform boosted trees on tabular data?
\newblock Advances in Neural Information Processing Systems 2023;\hspace{0pt}36:76336--76369.

\bibitem{shwartz2022tabular}
Shwartz-Ziv R, Armon A.
\newblock Tabular data: Deep learning is not all you need.
\newblock Information Fusion 2022;\hspace{0pt}81:84--90.
\newblock Doi: \url{https://doi.org/10.1016/j.inffus.2021.11.011}.

\bibitem{zabergja2024tabular}
Zab{\"e}rgja G, Kadra A, Grabocka J.
\newblock Tabular data: Is attention all you need?
\newblock arXiv preprint arXiv240203970 2024;\hspace{0pt}.

\bibitem{ott2024using}
Ott G, Schaubelt Y, Lopez~Alcaraz JM, Haverkamp W, Strodthoff N.
\newblock Using explainable ai to investigate electrocardiogram changes during healthy aging—from expert features to raw signals.
\newblock Plos one 2024;\hspace{0pt}19(4):e0302024.
\newblock Doi: \url{https://doi.org/10.1371/journal.pone.0302024}.

\bibitem{lundberg2020local}
Lundberg SM, Erion G, Chen H, DeGrave A, Prutkin JM, Nair B, Katz R, Himmelfarb J, Bansal N, Lee SI.
\newblock From local explanations to global understanding with explainable ai for trees.
\newblock Nature machine intelligence 2020;\hspace{0pt}2(1):56--67.
\newblock Doi: \url{https://doi.org/10.1038/s42256-019-0138-9}.

\bibitem{OuyangMICCAI23}
Ouyang D.
\newblock The uncanny and unreasonable performance of ai in medical imaging, 2023.
\newblock Doi: \url{https://sites.google.com/view/causemic}.

\bibitem{stuart2010matching}
Stuart EA.
\newblock Matching methods for causal inference: A review and a look forward.
\newblock Statistical science a review journal of the Institute of Mathematical Statistics 2010;\hspace{0pt}25(1):1.
\newblock Doi: \url{https://doi.org/10.1214/09-STS313}.

\bibitem{king2019propensity}
King G, Nielsen R.
\newblock Why propensity scores should not be used for matching.
\newblock Political analysis 2019;\hspace{0pt}27(4):435--454.
\newblock Doi: \url{https://doi.org/10.1017/pan.2019.11}.

\bibitem{sheedfar2013liver}
Sheedfar F, Biase SD, Koonen D, Vinciguerra M.
\newblock Liver diseases and aging: friends or foes?
\newblock Aging cell 2013;\hspace{0pt}12(6):950--954.
\newblock Doi: \url{https://doi.org/10.1111/acel.12128}.

\bibitem{hashimoto2011prevalence}
Hashimoto E, Tokushige K.
\newblock Prevalence, gender, ethnic variations, and prognosis of nash.
\newblock Journal of gastroenterology 2011;\hspace{0pt}46:63--69.
\newblock Doi: \url{https://doi.org/10.1007/s00535-010-0311-8}.

\bibitem{zhao2016prevalence}
Zhao J, Qi X, Hou F, Ning Z, Zhang X, Deng H, Peng Y, Li J, Wang X, Li H, et~al.
\newblock Prevalence, risk factors and in-hospital outcomes of qtc interval prolongation in liver cirrhosis.
\newblock The American journal of the medical sciences 2016;\hspace{0pt}352(3):285--295.
\newblock Doi: \url{https://doi.org/10.1016/j.amjms.2016.06.012}.

\bibitem{bal2003prolongation}
Bal JS, Thuluvath PJ.
\newblock Prolongation of qtc interval: relationship with etiology and severity of liver disease, mortality and liver transplantation.
\newblock Liver international 2003;\hspace{0pt}23(4):243--248.
\newblock Doi: \url{https://doi.org/10.1034/j.1600-0676.2003.00833.x}.

\bibitem{toma2020electrocardiographic}
Toma L, Stanciu AM, Zgura A, Bacalbasa N, Diaconu C, Iliescu L.
\newblock Electrocardiographic changes in liver cirrhosis—clues for cirrhotic cardiomyopathy.
\newblock Medicina 2020;\hspace{0pt}56(2):68.
\newblock Doi: \url{https://doi.org/10.3390/medicina56020068}.

\bibitem{lee2022prolonged}
Lee W, Vandenberk B, Raj SR, Lee SS.
\newblock Prolonged qt interval in cirrhosis: twisting time?
\newblock Gut and Liver 2022;\hspace{0pt}16(6):849.
\newblock Doi: \url{https://doi.org/10.5009/gnl210537}.

\bibitem{alcaraz2024causalconceptts}
Alcaraz JML, Strodthoff N.
\newblock Causalconceptts: Causal attributions for time series classification using high fidelity diffusion models.
\newblock arXiv preprint arXiv240515871 2024;\hspace{0pt}Doi: \url{https://arxiv.org/abs/2405.15871}.

\end{thebibliography}

\clearpage

\section*{Supplementary Material}

\subsection*{Additional results from the internal dataset}

\begin{table*}[h]
    \centering
    \begin{tabular}{lll}
        \hline
        \textbf{ICD-10 Code} & \textbf{AUROC (95\% CI)} & \textbf{Description} \\
        \hline
        \hline
        K860  & 0.9609 (0.9612, 0.9619) & Alcohol-induced chronic pancreatitis \\
        K852  & 0.9255 (0.9229, 0.9272) & Alcohol-induced acute pancreatitis \\
        K7011 & 0.9033 (0.9033, 0.9046) & Alcoholic hepatitis with ascites \\
        K7110 & 0.8784 (0.8791, 0.8807) & Toxic liver disease with hepatic necrosis, without coma \\
        K7010 & 0.8687 (0.8676, 0.8696) & Alcoholic hepatitis without ascites \\
        K761  & 0.8444 (0.8489, 0.8510) & Chronic passive congestion of liver \\
        K7031 & 0.8408 (0.8414, 0.8427) & Alcoholic cirrhosis of liver with ascites \\ 
        K7681 & 0.8380 (0.8420, 0.8435) & Hepatopulmonary syndrome \\
        K9423 & 0.8153 (0.8231, 0.8263) & Gastrostomy malfunction \\
        K652  & 0.7972 (0.7973, 0.8008) & Spontaneous bacterial peritonitis \\ 
        K767  & 0.7966 (0.7939, 0.7961) & Hepatorenal syndrome \\
        K704  & 0.7874 (0.7782, 0.7839) & Alcoholic hepatic failure \\ 
        K623  & 0.7840 (0.7741, 0.7836) & Rectal prolapse \\
        K8689 & 0.7699 (0.7771, 0.7819) & Other specified diseases of pancreas \\ 
        K8051 & 0.7671 (0.7655, 0.7684) & Calculus of bile duct without cholangitis or cholecystitis with obstruction \\
        K2920 & 0.7659 (0.7659, 0.7698) & Alcoholic gastritis without bleeding \\
        K7040 & 0.7600 (0.7522, 0.7557) & Alcoholic hepatic failure without coma \\
        K711  & 0.7462 (0.7470, 0.7544) & Toxic liver disease with hepatic necrosis \\
        K5732 & 0.7462 (0.7469, 0.7494) & Diverticulitis of large intestine without perforation or abscess without bleeding \\
        K807  & 0.7455 (0.7368, 0.7399) & Calculus of gallbladder and bile duct without cholecystitis \\
        K292  & 0.7450 (0.7476, 0.7510) & Alcoholic gastritis \\
        K5669 & 0.7442 (0.7437, 0.7481) & Other intestinal obstruction \\
        K3589 & 0.7421 (0.7416, 0.7462) & Other acute appendicitis \\
        K758  & 0.7381 (0.7346, 0.7406) & Other specified inflammatory liver diseases \\
        K264  & 0.7314 (0.7319, 0.7337) & Chronic or unspecified duodenal ulcer with hemorrhage \\
        K745  & 0.7313 (0.7307, 0.7350) & Biliary cirrhosis, unspecified \\
        K862  & 0.7287 (0.7211, 0.7277) & Cyst of pancreas \\
        K7291 & 0.7259 (0.7188, 0.7213) & Hepatic failure, unspecified with coma \\
        K7581 & 0.7212 (0.7245, 0.7291) & Nonalcoholic steatohepatitis (NASH) \\
        K2210 & 0.7129 (0.7007, 0.7161) & Ulcer of esophagus without bleeding \\
        K912  & 0.7099 (0.7113, 0.7154) & Postsurgical malabsorption, not elsewhere classified \\
        K762  & 0.7095 (0.7094, 0.7105) & Central hemorrhagic necrosis of liver \\
        K7200 & 0.7067 (0.7052, 0.7085) & Acute and subacute hepatic failure without coma \\
        K804  & 0.7050 (0.7005, 0.7033) & Calculus of bile duct with cholecystitis \\
        K029  & 0.7021 (0.7005, 0.7030) & Dental caries, unspecified \\
        K720  & 0.7009 (0.6996, 0.7034) & Acute and subacute hepatic failure \\
        \hline
    \end{tabular}
    \caption{Predictive performance across different ICD-10 codes from the internal dataset (MIMIC-IV-ECG), showing AUROC values for conditions that were not available in the external dataset (ECG-ViEW II).}
    \label{tab:icd10_codes}
\end{table*}

Table \ref{tab:icd10_codes} presents the predictive performance of our model across various ICD-10 codes from the internal dataset (MIMIC-IV-ECG), including conditions that were not available in the external validation dataset. Notably, the highest-performing conditions tend to be chronic hepatic diseases such as alcohol-induced chronic pancreatitis (K860, AUROC = 0.9609) and alcoholic hepatitis with ascites (K7011, AUROC = 0.9033). However, the table also includes several acute or subacute conditions, such as alcohol-induced acute pancreatitis (K852, AUROC = 0.9255) and acute and subacute hepatic failure (K720, AUROC = 0.7009), which allow for some assessment of the model's applicability beyond chronic diseases. While the model generally maintains good performance across a variety of conditions, the results highlight the need for future studies to explore additional non-alcoholic and acute hepatic conditions.

\subsection*{Model performance comparison for ECG features and demographics}

\begin{table*}[h]
    \centering
    \begin{tabular}{llll}
        \hline
        \textbf{ICD-10 Code} & \textbf{Diagnosis} & \textbf{Internal Test (AUC)} & \textbf{External Test (AUC)} \\
        \hline\hline
        \multicolumn{4}{c}{\textbf{ECG Features Only}} \\
        K70    & Alcoholic liver disease              & 0.7029 & 0.6423 \\
        K703   & Alcoholic cirrhosis                 & 0.7088 & 0.7753 \\
        K7030  & Alcoholic cirrhosis without ascites & 0.6753 & 0.7790 \\
        K729   & Hepatic failure, unspecified        & 0.7180 & 0.7500 \\
        K7290  & Hepatic failure without coma       & 0.7390 & 0.8149 \\
        K72    & Hepatic failure                     & 0.7002 & 0.7489 \\
        \hline
        \multicolumn{4}{c}{\textbf{Demographics Only}} \\
        K70    & Alcoholic liver disease              & 0.7352 & 0.7197 \\
        K703   & Alcoholic cirrhosis                 & 0.7157 & 0.7156 \\
        K7030  & Alcoholic cirrhosis without ascites & 0.7286 & 0.7119 \\
        K729   & Hepatic failure, unspecified        & 0.6645 & 0.6271 \\
        K7290  & Hepatic failure without coma       & 0.6497 & 0.6019 \\
        K72    & Hepatic failure                     & 0.6548 & 0.6006 \\
        \hline
        \multicolumn{4}{c}{\textbf{Combined (see main text)}} \\
        K70    & Alcoholic liver disease              & 0.8025 & 0.7644 \\
        K703   & Alcoholic cirrhosis                 & 0.7887 & 0.8590 \\
        K7030  & Alcoholic cirrhosis without ascites & 0.7819 & 0.8777 \\
        K729   & Hepatic failure, unspecified        & 0.7404 & 0.7498 \\
        K7290  & Hepatic failure without coma       & 0.7647 & 0.7821 \\
        K72    & Hepatic failure     &0.7833&0.8003\\
        \hline
    \end{tabular}
    \caption{Comparison of model performance (AUROC) for ECG-based and demographic-based predictions across internal and external test sets for different ICD-10 codes.}
    \label{tab:ecg_demographics}
\end{table*}

Table \ref{tab:ecg_demographics} presents the model performance (AUROC) for predicting different liver disease conditions using ECG features and demographic data. Notably, ECG-based models outperform demographic-based models in predicting hepatic failure (K72, K729, K7290), with higher AUROC scores in both internal and external test sets. This suggests that ECG signals capture relevant physiological changes associated with hepatic failure. Conversely, demographic-based models show stronger performance for alcoholic liver disease (K70) and alcoholic cirrhosis (K703, K7030), indicating that patient characteristics such as age, sex, and medical history play a more significant role in these conditions compared to ECG features. For comparison, we also indicate the results for the models operating on the combined feature set as presented in the main text. The combined models show statistically significant improved performance levels, assessed per non-overlapping confidence intervals, compared to single-modality models.

\subsection*{Model performance by demographic subgroups}

\begin{table*}[h]
    \centering
    \resizebox{\textwidth}{!}{%
    \begin{tabular}{lcccccc}
    \hline
    \textbf{Condition (ICD-10)} & \textbf{Female} & \textbf{Male} & \textbf{Q1 (18-53/18-40)} & \textbf{Q2 (53-66/40-52)} & \textbf{Q3 (66-78/52-65)} & \textbf{Q4 (78-101/65-109)} \\
    \hline\hline
    K70: Alcoholic liver disease & 0.8732/0.7528 & 0.7389/0.7060 & 0.8149/0.7320 & 0.7426/0.7610 & 0.7202/0.7395 & 0.3603/0.7480 \\
    K703: Alcoholic cirrhosis & 0.8538/0.8329 & 0.7234/0.8187 & 0.8059/0.9136 & 0.7206/0.8918 & 0.7153/0.8148 & 0.4422/0.7803 \\
    K7030: Alcoholic cirrhosis w/o ascites & 0.8481/0.9182 & 0.7105/0.8290 & 0.7996/0.9278 & 0.7571/0.9142 & 0.7046/0.8620 & 0.3546/0.8130 \\
    K72: Hepatic failure, NEC & 0.7017/0.7055 & 0.7523/0.7636 & 0.7798/0.6687 & 0.6443/0.7721 & 0.7044/0.7640 & 0.4660/0.6993 \\
    K729: Hepatic failure, unspecified & 0.7394/0.7276 & 0.7721/0.8045 & 0.8408/0.7213 & 0.6798/0.7995 & 0.6169/0.7819 & 0.4266/0.7070 \\
    K7290: Hepatic failure, unspecified w/o coma & 0.7647/0.7740 & 0.7841/0.8059 & 0.8568/0.6656 & 0.6328/0.8770 & 0.6968/0.7980 & 0.6569/0.7542 \\
    \hline
    \end{tabular}%
    }
    \caption{AUROC values for internal and external datasets, stratified by gender and age quantiles.}
    \label{tab:demographic_subgroups}
\end{table*}

Table \ref{tab:demographic_subgroups} presents the AUROC values for both internal and external datasets, stratified by gender and age quantiles. The results indicate consistent performance across different demographic groups, although certain subgroups exhibit variability, particularly in older age groups and female patients with specific conditions. The stratified AUROC values highlight potential demographic influences on model performance. In particular, performance varies across age quantiles, with a decrease in predictive accuracy for older patients in some conditions (e.g., K70, K703). This suggests that age-related physiological changes or variations in disease presentation may impact ECG-based predictions. Additionally, gender differences, such as higher AUROC for females in certain conditions (e.g., K7030), may reflect underlying biological or dataset distribution factors.

\subsection*{Hyperparameters}  

To ensure optimal model performance and prevent overfitting, we employed early stopping with a patience of 10 iterations, following standard practices for gradient-boosting models. Given that XGBoost inherently retains the best iteration based on validation performance, the total number of training iterations does not directly influence the final model.

We used the default hyperparameter settings of XGBoost without explicit tuning of the number of trees, maximum depth, or learning rate. However, we acknowledge that alternative hyperparameter configurations could impact model performance and may be explored in future work. The training hyperparameters for each model were as follows:   

\begin{itemize}
    \item \textbf{alpha}: 0  
    \item \textbf{cache\_opt}: 1  
    \item \textbf{colsample\_bylevel}: 1  
    \item \textbf{colsample\_bynode}: 1  
    \item \textbf{colsample\_bytree}: 1  
    \item \textbf{eta (learning rate)}: 0.300000012  
    \item \textbf{gamma}: 0  
    \item \textbf{grow\_policy}: depthwise  
    \item \textbf{interaction\_constraints}: None  
    \item \textbf{lambda (L2 regularization)}: 1  
    \item \textbf{max\_bin}: 256  
    \item \textbf{max\_cat\_threshold}: 64  
    \item \textbf{max\_cat\_to\_onehot}: 4  
    \item \textbf{max\_delta\_step}: 0  
    \item \textbf{max\_depth}: 6  
    \item \textbf{max\_leaves}: 0  
    \item \textbf{min\_child\_weight}: 1  
    \item \textbf{min\_split\_loss}: 0  
    \item \textbf{monotone\_constraints}: None  
    \item \textbf{refresh\_leaf}: 1  
    \item \textbf{reg\_alpha (L1 regularization)}: 0  
    \item \textbf{reg\_lambda (L2 regularization)}: 1  
    \item \textbf{sampling\_method}: uniform  
    \item \textbf{sketch\_ratio}: 2  
    \item \textbf{sparse\_threshold}: 0.2  
    \item \textbf{subsample}: 1  
    \item \textbf{num\_parallel\_tree}: 1  
    \item \textbf{num\_trees}: Between 25 and 37 per code/model  
\end{itemize}

\end{document}